\pdfoutput=1

\documentclass[11pt]{article}

\usepackage{emnlp2021}

\usepackage{times}
\usepackage{latexsym}

\usepackage[T1]{fontenc}

\usepackage[utf8]{inputenc}

\usepackage{microtype}

\usepackage{graphicx}
\usepackage{multirow}
\usepackage{multicol}
\usepackage{booktabs}

\title{Learning Hard Retrieval Decoder Attention for Transformers}

\author{Hongfei Xu$^1$\ \ \ \ Qiuhui Liu$^2$\ \ \ \ Josef van Genabith$^1$\thanks{\ \ \ \ Corresponding author.}\ \ \ \ Deyi Xiong$^{3,4}$\\
$^1$DFKI and Saarland University, Informatics Campus, Saarland, Germany\\
$^2$China Mobile Online Services, Henan, China\\
$^3$Tianjin University, Tianjin, China\\
$^4$Global Tone Communication Technology Co., Ltd.\\
\{hfxunlp, liuqhano\}@foxmail.com,
josef.van\_genabith@dfki.de,
dyxiong@tju.edu.cn\\}

\begin{document}
\maketitle
\begin{abstract}
The Transformer translation model is based on the multi-head attention mechanism, which can be parallelized easily. The multi-head attention network performs the scaled dot-product attention function in parallel, empowering the model by jointly attending to information from different representation subspaces at different positions. In this paper, we present an approach to learning a hard retrieval attention where an attention head only attends to one token in the sentence rather than all tokens. The matrix multiplication between attention probabilities and the value sequence in the standard scaled dot-product attention can thus be replaced by a simple and efficient retrieval operation. We show that our hard retrieval attention mechanism is $1.43$ times faster in decoding, while preserving translation quality on a wide range of machine translation tasks when used in the decoder self- and cross-attention networks.
\end{abstract}

\section{Introduction}

The Transformer translation model \citep{vaswani2017attention}, which has outperformed previous RNN/CNN based sequence-to-sequence models \citep{sutskever2014sequence,bahdanau2014neural,gehring2017convolutional}, is based on multi-head attention networks. The multi-head attention mechanism, which computes several scaled dot-product attentions in parallel, can be efficiently parallelized at sequence level.

In this paper, we investigate whether we can replace scaled dot-product attention by learning a simpler hard retrieval based attention that attends to a single token only. This simplifies computation and increases speed. We show that this can indeed be achieved by a simple and efficient retrieval operation while preserving translation quality.

Our contributions are as follows:

\begin{itemize}
	\item We propose a method to learn hard retrieval attention that attends with an efficient indexing operation (resulting in at most $h$ tokens being attended to for $h$ attention heads).
	\item We empirically show that using the hard retrieval attention mechanism for decoder self- and cross-attention networks increases the decoding speed by $1.43$ times while preserving performance on a wide range of MT tasks.
\end{itemize}

\section{Background: the Scaled Dot-Product Attention}

The multi-head attention network heavily employed by the Transformer translation model consists of $h$ parallel scaled dot-product attentions, where $h$ is the number of attention heads.

The scaled dot-product attention mechanism takes three inputs: the query sequence $Q$, the key sequence $K$ and the value sequence $V$.

It first compares each vector in $Q$ with all vectors in $K$ by dot-product computation to generate the attention score matrix $S$:

\begin{equation}
S = Q{K^T}
\label{eqa:comp}
\end{equation}

\noindent where $^T$ indicates matrix transposition.

Next, $S$ is scaled and normalized to attention probabilities $P$:

\begin{equation}
P = {\mathop{\rm softmax}\nolimits}(\frac{{S}}{{\sqrt {{d_k}} }})
\label{eqa:norm}
\end{equation}

\noindent where $d_k$ is the dimension of vectors of $K$.

Finally, the value sequence $V$ is weighted by the attention probabilities $P$ and accumulated as the attention result:

\begin{equation}
{\rm{Attention}}(Q,K,V) = PV
\label{eqa:attend}
\end{equation}

\section{Hard Retrieval Attention}

\subsection{Training}

\subsubsection{Forward Propagation}

When training the hard retrieval attention mechanism, we have to sharpen the attention probability vectors in $P$ (Eq. \ref{eqa:norm}) into one-hot vectors by multinomial sampling:

\begin{equation}
P_{hard} = {\rm{sharpen}}(P)
\label{eqa:sharp}
\end{equation}

\if false
We achieve this goal by sampling, where we denote $\overrightarrow p_j$ as the $j$th normalized vector of $P$, and $p_{j,i}$ as the probability that the $i$th element of $\overrightarrow p_j$ is turned to one. To strictly constrain that there is only one element that is turned into one and the others are all turned into zeros, we first perform a cumulative sum over $\overrightarrow p_j$:

\begin{equation}
\overrightarrow p_{cumsum\ j} = p_{j,1}, ..., \sum\limits_{x = 1}^i {{p_{j,x}}}, ..., \sum\limits_{x = 1}^n {{p_{j,x}}}
\label{eqa:cumsum}
\end{equation}

\noindent where $i$ and $n$ are the index of the $i$th element and the dimension of $\overrightarrow p_j$.

Next, we construct $n$ intervals based on $\overrightarrow p_{cumsum\ j}$:

\begin{equation}
[0, p_{j,1}], ..., [\sum\limits_{x = 1}^{n-1} {{p_{j,x}}}, \sum\limits_{x = 1}^n {{p_{j,x}}}]
\label{eqa:interval}
\end{equation}

Given these intervals, we sample a random number $k$ from the uniform distribution $U(0, 1]$. We select the interval of $k$ in Eq. \ref{eqa:interval}, and set the corresponding element to $1$ and the others to $0$ as the result of the sharpened one-hot vector $\overrightarrow p_{hard\ j}$.

We obtain $P_{hard}$ by applying the same process to all vectors of $P$. 
\fi

Since $P_{hard}$ only consists of one-hot vectors, the corresponding attention accumulation operation in Eq. \ref{eqa:attend} can be achieved efficiently by indexing. Specifically, for the $j$th one-hot vector $\overrightarrow p_{hard\ j}$, we first take the index $i_{max\ j}$ of the one-valued element:

\begin{equation}
i_{max\ j} = {\rm{argmax}}(\overrightarrow p_{hard\ j})
\label{eqa:argmax}
\end{equation}

\noindent where ${\rm{argmax}}$ returns the index of the largest element of a vector.

Note that in practice the multinomial sampling directly returns $i_{max\ j}$, but we keep $P_{hard}$ here to explain our approach in Section \ref{subsec:backward}.

Next, we obtain the hard attention result with a simple retrieval operation on $V$:

\begin{equation}
{\rm{Attention}}_{Hard}(Q,K,V)[j] = V[i_{max\ j}]
\label{eqa:hardattend}
\end{equation}

\subsubsection{Gradient Computation}
\label{subsec:backward}

To compute gradients for the hard attention mechanism, we address the non-differentiability of the sharpening operation in the forward propagation by regarding it as a noise process:

\begin{equation}
P_{hard} = P + Noise
\label{eqa:noise}
\end{equation}

\noindent where $Noise$ stands for the noise introduced by the sharpening operation.

Thus, we pass the gradients $P_{hard}^g$ of $P_{hard}$ directly to $P$ to fix the chain rule for back propagation:

\begin{equation}
P^g = P_{hard}^g
\label{eqa:gradp}
\end{equation}

\noindent where $P^g$ stands for the gradient of $P$.

The retrieval operation of the value sequence $V$ with the matrix $P_{hard}$ consisting of one-hot vectors is equivalent to the matrix-multiplication between $P_{hard}$ and $V$. Given the gradient of the hard attention result $Attention_{Hard}^g$, the gradients of $P_{hard}$ and $V$ can be computed as:

\begin{equation}
P_{hard}^g = Attention_{Hard}^gV^T
\label{eqa:gradphard}
\end{equation}

\begin{equation}
{V^g}[i] = \left\{ {\begin{array}{*{20}{c}}
{\sum {Attention_{Hard}^g[j]} ,\ }&{i = {i_{max\ j}}}\\
0,\ &{{\rm{otherwise}}}
\end{array}} \right.
\label{eqa:gradV}
\end{equation}

\noindent where $V^g[i]$ is the $i$th row of the gradient matrix $V^g$ of $V$.

For efficiency, we use the retrieval operation again instead of the matrix multiplication for the computation of $V^g$ like in the forward pass.

\subsection{Inference}

Since the largest attention score in Eq. \ref{eqa:comp} corresponds to the largest probability after scaling and normalization in Eq. \ref{eqa:norm}, we skip the computation of Eq. \ref{eqa:norm} and directly take the result of Eq. \ref{eqa:comp} for the computation of retrieval indexes during inference:

\begin{equation}
i_{max\ j} = {\rm{argmax}}(\overrightarrow s_{j})
\label{eqa:argmaxeval}
\end{equation}

\noindent where $\overrightarrow s_{j}$ stands for the $j$th row of $S$.

Next, we can obtain the hard attention results with $i_{max\ j}$ by the simple retrieval operation presented in Eq. \ref{eqa:hardattend}.

\if false
\begin{equation}
{\rm{Attention}}_{Hard}(Q,K,V)[j] = V[i_{max\ j}]
\label{eqa:hardattendeval}
\end{equation}
\fi

\section{Experiment}

We implemented our approach based on the Neutron implementation of the Transformer \citep{xu2019neutron}.

To investigate the impact on translation quality of our approach, we conducted our experiments on the WMT 14 English to German and English to French news translation tasks to compare with \citet{vaswani2017attention}. We also examined the impact of our approach on the pre-processed data of the WMT 17 news translation tasks for $12$ translation directions.

The concatenation of newstest 2012 and newstest 2013 was used for validation and newstest 2014 as test sets for the WMT 14 English to German and English to French news translation tasks. We used the pre-processed data for WMT 17 news translation tasks.\footnote{\url{http://data.statmt.org/wmt17/translation-task/preprocessed/}.}

\subsection{Settings}

We applied joint Byte-Pair Encoding (BPE) \citep{sennrich2015neural} with $32k$ merging operations on both data sets to address the unknown word issue. We only kept sentences with a maximum of $256$ subword tokens for training. Training sets were randomly shuffled in every training epoch. We followed \citet{vaswani2017attention} for experiment settings. We used a beam size of $4$ for decoding with the averaged model of the last $5$ checkpoints for the Transformer Base setting and $20$ checkpoints for the Transformer Big setting saved with an interval of $1,500$ training steps, and evaluated tokenized case-sensitive BLEU.

Though \citet{zhang2019improving,xu2020dynamically} suggest using a large batch size which may lead to improved performance, we used a batch size of $25k$ target tokens which was achieved through gradient accumulation of small batches to fairly compare with \citet{vaswani2017attention}. The training steps for Transformer Base and Transformer Big were $100k$ and $300k$ respectively following \citet{vaswani2017attention}. Parameters were initialized under the Lipschitz constraint \citep{xu2020lipschitz}.

\begin{table}[t]
  \centering
    \begin{tabular}{lrr}
    \toprule
    Models & \multicolumn{1}{l}{En-De} & \multicolumn{1}{l}{En-Fr} \\
    \midrule
    Transformer Base & 27.55 & 39.54 \\
    with Hard Dec Attn & 27.73 & 39.39 \\
    \midrule
    Transformer Big & 28.63 & 41.52 \\
    with Hard Dec Attn & 28.42 & 41.81 \\
    \bottomrule
    \end{tabular}%
  \caption{Results on WMT 14 En-De and En-Fr.}
  \label{tab:bleumain}%
\end{table}%

\begin{table*}[t]
  \centering
    \begin{tabular}{ccrr}
    \toprule
    \multicolumn{2}{c}{Operation} & \multicolumn{2}{c}{Costs} \\
    \multicolumn{1}{c}{Std} & \multicolumn{1}{c}{Hard} & \multicolumn{1}{c}{Std} & \multicolumn{1}{c}{Hard} \\
    \midrule
    \multicolumn{2}{c}{Compare (Eq. \ref{eqa:comp})} & \multicolumn{2}{c}{14.40} \\
    \multicolumn{1}{c}{Normalize (Eq. \ref{eqa:norm})} & argmax (Eq. \ref{eqa:argmaxeval}) & 6.22  & 2.10  \\
    \multicolumn{1}{c}{Attend (Eq. \ref{eqa:attend})} & Index (Eq. \ref{eqa:hardattend}) & 12.26 & 2.93 \\
    \midrule
    \multicolumn{2}{c}{Total} & 32.88  & 19.43  \\
    \bottomrule
    \end{tabular}%
  \caption{Time costs (in seconds) of operations for 1000 iterations. Std: standard scaled dot-product attention.}
  \label{tab:effrs}%
\end{table*}%

\begin{table*}[t]
	\centering
	\begin{tabular}{lrrr}
		\toprule
		Hard Attention & \multicolumn{1}{c}{BLEU} & \multicolumn{1}{c}{Decoding Speed (sent/s)}  & \multicolumn{1}{c}{Speed-Up}\\
		\midrule
		None  & 27.55 & 150.15 & 1.00 \\
		Dec Cross-Attn & 27.59 & 187.69 & 1.25 \\
		Dec Cross- and Self-Attn & 27.73 & 214.50 & 1.43 \\
		Enc Self- and Dec Cross- and Self-Attn & 26.60 & 219.20 & 1.46 \\
		\bottomrule
	\end{tabular}%
	\caption{Results on using hard attention for different attention mechanisms. Speed is measured on the WMT 14 En-De testset with a beam size of $4$.}
	\label{tab:abl}%
\end{table*}%

\subsection{Main Results}

We first examine the effects of using hard retrieval attention for decoder self- and cross-attention networks (reported in our ablation study results in Table \ref{tab:abl}) on the WMT 14 English-German and English-French task to compare with \citet{vaswani2017attention}. Results are shown in Table \ref{tab:bleumain}.

Table \ref{tab:bleumain} shows that using the hard retrieval attention mechanism for decoder self- and cross-attention networks achieves comparable performance on both tasks under both Transformer Base and Big settings.

\subsection{Efficiency Analysis}

Comparing the inference of the standard scaled dot-product attention with the hard retrieval attention, we expect the latter to be faster and more efficient than the first as:

\begin{itemize}
	\item The operation to find the index of the largest element in the vector (Eq. \ref{eqa:argmaxeval}) in the hard retrieval attention is more efficient than the scaling and normalization (Eq. \ref{eqa:norm}) in the scaled dot-product attention.
	\item The operation to retrieve the corresponding vector in $V$ with indexes in the hard retrieval attention is faster than the matrix multiplication in the standard attention.
\end{itemize}

We tested the efficiency of our approach by recording the time cost of the operations involved in the two attention mechanisms during the forward propagation on the development set of the WMT 14 English-German news translation task with a single GTX 1080 Ti GPU under the Transformer Base setting. Results are shown in Table \ref{tab:effrs}. Table \ref{tab:effrs} shows that our hard retrieval attention is much faster than scaled dot-product attention.

Even though overall time consumption is not only determined by attention networks, but also by other parts of the Transformer, we suggest the acceleration with hard retrival attention during inference is still significant, as decoding is performed autoregressively in a token-by-token manner and decoder layers have to be computed for many times during inference, while the linear projections for keys and values of the decoder self-attention and cross-attention heads will be computed once only and cached. This makes attention computation consume a larger part of computation during inference than during training, and makes the acceleration of decoder attention layers significant. Using hard attention also saves the computation of the linear projection layer for values, as it only needs to compute the representations of several attended tokens instead of all tokens of the sequence. We report overall decoding speed in Table \ref{tab:abl}.

\begin{table}[t]
  \centering
    \begin{tabular}{ccr}
    \toprule
    Train & Decode & \multicolumn{1}{c}{BLEU} \\
    \midrule
	\multicolumn{2}{c}{Soft} & 27.55 \\
    Soft  & Hard  & 26.30 \\
    \multicolumn{2}{c}{Hard} & 27.73 \\
    Hard  & Soft  & 27.80 \\
    \bottomrule
    \end{tabular}%
  \caption{Effects of hard attention training and decoding on WMT 14 En-De.}
  \label{tab:shlearn}%
\end{table}%

\begin{table*}[t]
  \centering
    \begin{tabular}{lrrrrr}
    \toprule
    \multicolumn{1}{c}{\multirow{2}[0]{*}{Lang}} & \multicolumn{1}{c}{\multirow{2}[0]{*}{Data (M)}} & \multicolumn{2}{c}{En$\rightarrow$xx} & \multicolumn{2}{c}{xx$\rightarrow$En} \\
          &       & \multicolumn{1}{c}{Std} & \multicolumn{1}{c}{Hard} & \multicolumn{1}{c}{Std} & \multicolumn{1}{c}{Hard} \\
    \midrule
    De    & 5.85  & 27.48  & 27.56  & 32.89  & 32.68  \\
    Fi    & 2.63  & 22.23  & 22.15  & 26.15  & 26.03  \\
    Lv    & 4.46  & 16.38  & 16.43  & 18.12  & 18.20  \\
    Ru    & 25.00  & 28.20  & 28.04  & 31.52  & 31.30  \\
    Tr    & 0.20  & 15.79  & 15.81  & 15.58  & 15.69  \\
    Cs    & 52.02  & 21.89  & 21.78  & 27.62  & 27.36  \\
	\midrule
    Avg.    &  & 22.00  & 21.96  & 25.31  & 25.20  \\
    \bottomrule
    \end{tabular}%
  \caption{Results on WMT 17 news translation tasks. xx denotes the language in row headers. None of the differences are statistically significant.}
  \label{tab:verify}%
\end{table*}%

\subsection{Ablation Study}
\label{subsec:abl}

We conducted ablation studies on the WMT 14 En-De task.

We first test decoding with the hard retrival attention algorithm but with the converged standard Transformer model. Results are shown in Table \ref{tab:shlearn}.

Table \ref{tab:shlearn} shows that performing hard decoding with the softly trained model leads to significant loss in BLEU ($-1.21$). On the one hand this shows that the decoder attention network does not really need to attend too many tokens, on the other hand it shows the importance of our hard attention training approach that closes the gap between training and inference.

We also study applying the hard retrival attention network as different attention sub-layers of the decoder or both encoder and decoder. Results are shown in Table \ref{tab:abl}.

Table \ref{tab:abl} shows that applying the hard retrieval attention mechanism to encoder self-attention networks significantly hampers performance. We conjecture potential reasons might be: 1) the encoder might be harder to train than the decoder as its gradients come from cross-attention networks while the decoder receives more direct supervision from the classifier, and the hard attention training approach makes the encoder's training even harder. 2) as the hard retrieval attention only attends one token, the multi-head hard retrieval attention can only attend at most the same number of tokens as the number of attention heads. To achieve optimal results, encoder self-attention may need to attend to more tokens. We leave how to use hard retrieval attention in the encoder for future work. Fortunately, the autoregressive decoder is the major factor in time consumption during decoding, and accelerating decoder layers' computation can significantly speed up inference.

\subsection{Testing on WMT 17 Tasks}

We further examine the performance of using hard retrieval attention for decoder attention networks on all WMT 17 news translation tasks, using the same setting of the Transformer Base as on the WMT 14 En-De task. Results are shown in Table \ref{tab:verify}. Table \ref{tab:verify} shows that hard retrieval attention is able to match the performance in all tested language pairs in both translation directions, with training sets ranging from $0.2$M to $52.02$M sentence pairs. The largest performance loss ($-0.26$ BLEU) is on the Cs-En task.

\section{Related Work}

\citet{zhang2018accelerating} accelerate the decoder self-attention with the average attention network. \citet{xu2021multi} propose to replace the self-attention layer by multi-head highly parallelized LSTM. \citet{kim2019research} investigate knowledge distillation and quantization for faster NMT decoding. \citet{yi2020synthesizer} investigate the true importance and contribution of the dot product-based self-attention mechanism on the performance of Transformer models.

Most previous research focuses on efficient modeling of the self-attention mechanism for very long sequences. These are generally not effective on sequences of normal lengths. \citet{dai2019transformer} introduce the notion of recurrence into deep self-attention network to model very long term dependency efficiently. \citet{ma2019tensorized} combine low rank approximate and parameter sharing to construct a tensorized Transformer. \citet{kitaev2020reformer} replace dot-product attention by one that uses locality-sensitive hashing and use reversible residual layers instead of the standard residuals. \citet{zhang2020tensorcoder} propose a dimension-wise attention mechanism to reduce the attention complexity. \citet{angelos2020transformers} express the self-attention as a linear dot-product of kernel feature maps and make use of the associativity property of matrix products. \citet{wang2020linformer} approximate the self-attention mechanism by a low-rank matrix. \citet{iz2020longformer} introduce an attention mechanism that scales linearly with sequence length. \citet{rewon2020generating} introduce sparse factorizations of the attention matrix.

On using hard (local) attention for machine translation, \citet{luong2015effective} selectively focus on a small window of context smoothed by a Gaussian distribution. For self-attentional sentence encoding, \citet{shen2018reinforced} train hard attention mechanisms which select a subset of tokens via policy gradient. \citet{geng2020selective} investigate selective self-attention networks implemented with Gumble-Sigmoid. Sparse attention has been found benefitial for performance \citep{malaviya2018sparse,peters2019sparse,correia2019adaptively,indurthi2019look,maruf2019selective}. Our approach learns explicit one-to-one attention for efficiency, pushing such research efforts to the limit.

\section{Conclusion}

We propose to learn a hard retrieval attention which only attends to one token rather than all tokens. With the one-to-one hard attention matrix, the matrix multiplication between attention probabilities and the value sequence in the standard scaled dot-product attention can be replaced by a simple and efficient retrieval operation.

In our experiments on a wide range of machine translation tasks, we show that using the hard retrieval attention for decoder attention networks can achieve competitive performance while being $1.43$ times faster in decoding.

\section*{Acknowledgements}

We thank our anonymous reviewers for their insightful comments. Hongfei Xu acknowledges the support of China Scholarship Council ([2018]3101, 201807040056). Josef van Genabith and Hongfei Xu are supported by the German Federal Ministry of Education and Research (BMBF) under funding code 01IW20010 (CORA4NLP). Deyi Xiong is partially supported by Natural Science Foundation of Tianjin (Grant No. 19JCZDJC31400) and the joint research center between GTCOM and Tianjin University.

\bibliography{custom}

\begin{thebibliography}{29}
\expandafter\ifx\csname natexlab\endcsname\relax\def\natexlab#1{#1}\fi

\bibitem[{Bahdanau et~al.(2015)Bahdanau, Cho, and Bengio}]{bahdanau2014neural}
Dzmitry Bahdanau, Kyunghyun Cho, and Yoshua Bengio. 2015.
\newblock \href {http://arxiv.org/abs/1409.0473} {Neural machine translation by
  jointly learning to align and translate}.
\newblock In \emph{3rd International Conference on Learning Representations,
  {ICLR} 2015, San Diego, CA, USA, May 7-9, 2015, Conference Track
  Proceedings}.

\bibitem[{Beltagy et~al.(2020)Beltagy, Peters, and Cohan}]{iz2020longformer}
Iz~Beltagy, Matthew~E. Peters, and Arman Cohan. 2020.
\newblock \href {http://arxiv.org/abs/2004.05150} {Longformer: The
  long-document transformer}.
\newblock \emph{CoRR}, abs/2004.05150.

\bibitem[{Child et~al.(2019)Child, Gray, Radford, and
  Sutskever}]{rewon2020generating}
Rewon Child, Scott Gray, Alec Radford, and Ilya Sutskever. 2019.
\newblock \href {http://arxiv.org/abs/1904.10509} {Generating long sequences
  with sparse transformers}.
\newblock \emph{CoRR}, abs/1904.10509.

\bibitem[{Correia et~al.(2019)Correia, Niculae, and
  Martins}]{correia2019adaptively}
Gon{\c{c}}alo~M. Correia, Vlad Niculae, and Andr{\'e} F.~T. Martins. 2019.
\newblock \href {https://doi.org/10.18653/v1/D19-1223} {Adaptively sparse
  transformers}.
\newblock In \emph{Proceedings of the 2019 Conference on Empirical Methods in
  Natural Language Processing and the 9th International Joint Conference on
  Natural Language Processing (EMNLP-IJCNLP)}, pages 2174--2184, Hong Kong,
  China. Association for Computational Linguistics.

\bibitem[{Dai et~al.(2019)Dai, Yang, Yang, Carbonell, Le, and
  Salakhutdinov}]{dai2019transformer}
Zihang Dai, Zhilin Yang, Yiming Yang, Jaime Carbonell, Quoc Le, and Ruslan
  Salakhutdinov. 2019.
\newblock \href {https://doi.org/10.18653/v1/P19-1285} {Transformer-{XL}:
  Attentive language models beyond a fixed-length context}.
\newblock In \emph{Proceedings of the 57th Annual Meeting of the Association
  for Computational Linguistics}, pages 2978--2988, Florence, Italy.
  Association for Computational Linguistics.

\bibitem[{Gehring et~al.(2017)Gehring, Auli, Grangier, Yarats, and
  Dauphin}]{gehring2017convolutional}
Jonas Gehring, Michael Auli, David Grangier, Denis Yarats, and Yann~N. Dauphin.
  2017.
\newblock \href {http://proceedings.mlr.press/v70/gehring17a.html}
  {Convolutional sequence to sequence learning}.
\newblock In \emph{Proceedings of the 34th International Conference on Machine
  Learning}, volume~70 of \emph{Proceedings of Machine Learning Research},
  pages 1243--1252, International Convention Centre, Sydney, Australia. PMLR.

\bibitem[{Geng et~al.(2020)Geng, Wang, Wang, Qin, Liu, and
  Tu}]{geng2020selective}
Xinwei Geng, Longyue Wang, Xing Wang, Bing Qin, Ting Liu, and Zhaopeng Tu.
  2020.
\newblock \href {https://doi.org/10.18653/v1/2020.acl-main.269} {How does
  selective mechanism improve self-attention networks?}
\newblock In \emph{Proceedings of the 58th Annual Meeting of the Association
  for Computational Linguistics}, pages 2986--2995, Online. Association for
  Computational Linguistics.

\bibitem[{Indurthi et~al.(2019)Indurthi, Chung, and Kim}]{indurthi2019look}
Sathish~Reddy Indurthi, Insoo Chung, and Sangha Kim. 2019.
\newblock \href {https://doi.org/10.18653/v1/P19-1290} {Look harder: A neural
  machine translation model with hard attention}.
\newblock In \emph{Proceedings of the 57th Annual Meeting of the Association
  for Computational Linguistics}, pages 3037--3043, Florence, Italy.
  Association for Computational Linguistics.

\bibitem[{Katharopoulos et~al.(2020)Katharopoulos, Vyas, Pappas, and
  Fleuret}]{angelos2020transformers}
Angelos Katharopoulos, Apoorv Vyas, Nikolaos Pappas, and Fran{\c{c}}ois
  Fleuret. 2020.
\newblock \href {https://proceedings.mlr.press/v119/katharopoulos20a.html}
  {Transformers are {RNN}s: Fast autoregressive transformers with linear
  attention}.
\newblock In \emph{Proceedings of the 37th International Conference on Machine
  Learning}, volume 119 of \emph{Proceedings of Machine Learning Research},
  pages 5156--5165. PMLR.

\bibitem[{Kim et~al.(2019)Kim, Junczys-Dowmunt, Hassan, Fikri~Aji, Heafield,
  Grundkiewicz, and Bogoychev}]{kim2019research}
Young~Jin Kim, Marcin Junczys-Dowmunt, Hany Hassan, Alham Fikri~Aji, Kenneth
  Heafield, Roman Grundkiewicz, and Nikolay Bogoychev. 2019.
\newblock \href {https://doi.org/10.18653/v1/D19-5632} {From research to
  production and back: Ludicrously fast neural machine translation}.
\newblock In \emph{Proceedings of the 3rd Workshop on Neural Generation and
  Translation}, pages 280--288, Hong Kong. Association for Computational
  Linguistics.

\bibitem[{Kitaev et~al.(2020)Kitaev, Kaiser, and Levskaya}]{kitaev2020reformer}
Nikita Kitaev, Lukasz Kaiser, and Anselm Levskaya. 2020.
\newblock \href {https://openreview.net/forum?id=rkgNKkHtvB} {Reformer: The
  efficient transformer}.
\newblock In \emph{International Conference on Learning Representations}.

\bibitem[{Luong et~al.(2015)Luong, Pham, and Manning}]{luong2015effective}
Thang Luong, Hieu Pham, and Christopher~D. Manning. 2015.
\newblock \href {https://doi.org/10.18653/v1/D15-1166} {Effective approaches to
  attention-based neural machine translation}.
\newblock In \emph{Proceedings of the 2015 Conference on Empirical Methods in
  Natural Language Processing}, pages 1412--1421, Lisbon, Portugal. Association
  for Computational Linguistics.

\bibitem[{Ma et~al.(2019)Ma, Zhang, Zhang, Duan, Hou, Zhou, and
  Song}]{ma2019tensorized}
Xindian Ma, Peng Zhang, Shuai Zhang, Nan Duan, Yuexian Hou, Ming Zhou, and
  Dawei Song. 2019.
\newblock \href
  {http://papers.nips.cc/paper/8495-a-tensorized-transformer-for-language-modeling.pdf}
  {A tensorized transformer for language modeling}.
\newblock In H.~Wallach, H.~Larochelle, A.~Beygelzimer, F.~d\textquotesingle
  Alch\'{e}-Buc, E.~Fox, and R.~Garnett, editors, \emph{Advances in Neural
  Information Processing Systems 32}, pages 2232--2242. Curran Associates, Inc.

\bibitem[{Malaviya et~al.(2018)Malaviya, Ferreira, and
  Martins}]{malaviya2018sparse}
Chaitanya Malaviya, Pedro Ferreira, and Andr{\'e} F.~T. Martins. 2018.
\newblock \href {https://doi.org/10.18653/v1/P18-2059} {Sparse and constrained
  attention for neural machine translation}.
\newblock In \emph{Proceedings of the 56th Annual Meeting of the Association
  for Computational Linguistics (Volume 2: Short Papers)}, pages 370--376,
  Melbourne, Australia. Association for Computational Linguistics.

\bibitem[{Maruf et~al.(2019)Maruf, Martins, and Haffari}]{maruf2019selective}
Sameen Maruf, Andr{\'e} F.~T. Martins, and Gholamreza Haffari. 2019.
\newblock \href {https://doi.org/10.18653/v1/N19-1313} {Selective attention for
  context-aware neural machine translation}.
\newblock In \emph{Proceedings of the 2019 Conference of the North {A}merican
  Chapter of the Association for Computational Linguistics: Human Language
  Technologies, Volume 1 (Long and Short Papers)}, pages 3092--3102,
  Minneapolis, Minnesota. Association for Computational Linguistics.

\bibitem[{Peters et~al.(2019)Peters, Niculae, and Martins}]{peters2019sparse}
Ben Peters, Vlad Niculae, and Andr{\'e} F.~T. Martins. 2019.
\newblock \href {https://doi.org/10.18653/v1/P19-1146} {Sparse
  sequence-to-sequence models}.
\newblock In \emph{Proceedings of the 57th Annual Meeting of the Association
  for Computational Linguistics}, pages 1504--1519, Florence, Italy.
  Association for Computational Linguistics.

\bibitem[{Sennrich et~al.(2016)Sennrich, Haddow, and
  Birch}]{sennrich2015neural}
Rico Sennrich, Barry Haddow, and Alexandra Birch. 2016.
\newblock \href {https://doi.org/10.18653/v1/P16-1162} {Neural machine
  translation of rare words with subword units}.
\newblock In \emph{Proceedings of the 54th Annual Meeting of the Association
  for Computational Linguistics (Volume 1: Long Papers)}, pages 1715--1725.
  Association for Computational Linguistics.

\bibitem[{Shen et~al.(2018)Shen, Zhou, Long, Jiang, Wang, and
  Zhang}]{shen2018reinforced}
Tao Shen, Tianyi Zhou, Guodong Long, Jing Jiang, Sen Wang, and Chengqi Zhang.
  2018.
\newblock \href {https://doi.org/10.24963/ijcai.2018/604} {Reinforced
  self-attention network: a hybrid of hard and soft attention for sequence
  modeling}.
\newblock In \emph{Proceedings of the Twenty-Seventh International Joint
  Conference on Artificial Intelligence, {IJCAI-18}}, pages 4345--4352.
  International Joint Conferences on Artificial Intelligence Organization.

\bibitem[{Sutskever et~al.(2014)Sutskever, Vinyals, and
  Le}]{sutskever2014sequence}
Ilya Sutskever, Oriol Vinyals, and Quoc~V Le. 2014.
\newblock \href
  {http://papers.nips.cc/paper/5346-sequence-to-sequence-learning-with-neural-networks.pdf}
  {Sequence to sequence learning with neural networks}.
\newblock In Z.~Ghahramani, M.~Welling, C.~Cortes, N.~D. Lawrence, and K.~Q.
  Weinberger, editors, \emph{Advances in Neural Information Processing Systems
  27}, pages 3104--3112. Curran Associates, Inc.

\bibitem[{Tay et~al.(2021)Tay, Bahri, Metzler, Juan, Zhao, and
  Zheng}]{yi2020synthesizer}
Yi~Tay, Dara Bahri, Donald Metzler, Da-Cheng Juan, Zhe Zhao, and Che Zheng.
  2021.
\newblock \href {https://openreview.net/forum?id=H-SPvQtMwm} {Synthesizer:
  Rethinking self-attention for transformer models}.

\bibitem[{Vaswani et~al.(2017)Vaswani, Shazeer, Parmar, Uszkoreit, Jones,
  Gomez, Kaiser, and Polosukhin}]{vaswani2017attention}
Ashish Vaswani, Noam Shazeer, Niki Parmar, Jakob Uszkoreit, Llion Jones,
  Aidan~N Gomez, \L~ukasz Kaiser, and Illia Polosukhin. 2017.
\newblock \href
  {http://papers.nips.cc/paper/7181-attention-is-all-you-need.pdf} {Attention
  is all you need}.
\newblock In \emph{Advances in Neural Information Processing Systems 30}, pages
  5998--6008. Curran Associates, Inc.

\bibitem[{Wang et~al.(2020)Wang, Li, Khabsa, Fang, and Ma}]{wang2020linformer}
Sinong Wang, Belinda~Z. Li, Madian Khabsa, Han Fang, and Hao Ma. 2020.
\newblock \href {http://arxiv.org/abs/2006.04768} {Linformer: Self-attention
  with linear complexity}.

\bibitem[{Xu and Liu(2019)}]{xu2019neutron}
Hongfei Xu and Qiuhui Liu. 2019.
\newblock \href {http://arxiv.org/abs/1903.07402} {{Neutron: An Implementation
  of the Transformer Translation Model and its Variants}}.
\newblock \emph{arXiv preprint arXiv:1903.07402}.

\bibitem[{Xu et~al.(2020{\natexlab{a}})Xu, Liu, van Genabith, Xiong, and
  Zhang}]{xu2020lipschitz}
Hongfei Xu, Qiuhui Liu, Josef van Genabith, Deyi Xiong, and Jingyi Zhang.
  2020{\natexlab{a}}.
\newblock \href {https://www.aclweb.org/anthology/2020.acl-main.38} {Lipschitz
  constrained parameter initialization for deep transformers}.
\newblock In \emph{Proceedings of the 58th Annual Meeting of the Association
  for Computational Linguistics}, pages 397--402, Online. Association for
  Computational Linguistics.

\bibitem[{Xu et~al.(2021)Xu, Liu, van Genabith, Xiong, and Zhang}]{xu2021multi}
Hongfei Xu, Qiuhui Liu, Josef van Genabith, Deyi Xiong, and Meng Zhang. 2021.
\newblock \href {https://doi.org/10.18653/v1/2021.acl-long.23} {Multi-head
  highly parallelized {LSTM} decoder for neural machine translation}.
\newblock In \emph{Proceedings of the 59th Annual Meeting of the Association
  for Computational Linguistics and the 11th International Joint Conference on
  Natural Language Processing (Volume 1: Long Papers)}, pages 273--282, Online.
  Association for Computational Linguistics.

\bibitem[{Xu et~al.(2020{\natexlab{b}})Xu, van Genabith, Xiong, and
  Liu}]{xu2020dynamically}
Hongfei Xu, Josef van Genabith, Deyi Xiong, and Qiuhui Liu. 2020{\natexlab{b}}.
\newblock \href {https://www.aclweb.org/anthology/2020.acl-main.323}
  {Dynamically adjusting transformer batch size by monitoring gradient
  direction change}.
\newblock In \emph{Proceedings of the 58th Annual Meeting of the Association
  for Computational Linguistics}, pages 3519--3524, Online. Association for
  Computational Linguistics.

\bibitem[{Zhang et~al.(2019)Zhang, Titov, and Sennrich}]{zhang2019improving}
Biao Zhang, Ivan Titov, and Rico Sennrich. 2019.
\newblock \href {https://doi.org/10.18653/v1/D19-1083} {Improving deep
  transformer with depth-scaled initialization and merged attention}.
\newblock In \emph{Proceedings of the 2019 Conference on Empirical Methods in
  Natural Language Processing and the 9th International Joint Conference on
  Natural Language Processing (EMNLP-IJCNLP)}, pages 898--909, Hong Kong,
  China. Association for Computational Linguistics.

\bibitem[{Zhang et~al.(2018)Zhang, Xiong, and Su}]{zhang2018accelerating}
Biao Zhang, Deyi Xiong, and Jinsong Su. 2018.
\newblock \href {https://doi.org/10.18653/v1/P18-1166} {Accelerating neural
  transformer via an average attention network}.
\newblock In \emph{Proceedings of the 56th Annual Meeting of the Association
  for Computational Linguistics (Volume 1: Long Papers)}, pages 1789--1798,
  Melbourne, Australia. Association for Computational Linguistics.

\bibitem[{Zhang et~al.(2020)Zhang, Zhang, Ma, Wei, Wang, and
  Liu}]{zhang2020tensorcoder}
Shuai Zhang, Peng Zhang, Xindian Ma, Junqiu Wei, Ningning Wang, and Qun Liu.
  2020.
\newblock \href {http://arxiv.org/abs/2008.01547} {Tensorcoder: Dimension-wise
  attention via tensor representation for natural language modeling}.
\newblock \emph{CoRR}, abs/2008.01547.

\end{thebibliography}
\bibliographystyle{acl_natbib}

\end{document}